\documentclass[times,twocolumn,final]{elsarticle}
\usepackage{framed,multirow}

\usepackage{amssymb}
\usepackage{latexsym}
\usepackage{hyperref}       
\usepackage{url}            
\usepackage{booktabs}       
\usepackage{amsfonts}
\usepackage{nicefrac}       
\usepackage{microtype}     
\usepackage{xcolor}   
\usepackage{graphicx}
\usepackage{footnote}
\usepackage{adjustbox}
\usepackage{pifont} 
\usepackage{subfigure}
\usepackage{xcolor} 
\usepackage{pifont}

\usepackage{pifont}

\usepackage{url}
\usepackage{xcolor}
\usepackage{amsmath}
\usepackage{hyperref}
\usepackage{tabularx,colortbl}

\definecolor{newcolor}{rgb}{.8,.349,.1}

\begin{document}


\begin{frontmatter}

\title{L2GNet: Optimal Local-to-Global Representation of Anatomical Structures for Generalized Medical Image Segmentation}%

\author[1]{Vandan Gorade}
\author[2]{Sparsh Mittal}
\author[3]{Neethi Dasu}
\author[4]{Rekha Singhal}
\author[5]{KC Santosh}
\author[5]{Debesh Jha}

\newcommand{\KWD}[1]{\textbf{Keywords:} #1}

\cortext[cor1]{Corresponding author: debesh.jha@usd.edu. This work is supported by the IIT Roorkee grant FIG-100874.}

\address[1]{Department of Biomedical Engineering, Northwestern University, IL, USA}
\address[2]{Mehta School of Data Science and AI, Indian Institute of Technology, Roorkee, India}
\address[3]{Beth Israel Lahey Health, University of Massachusetts Chan School of Medicine, USA}
\address[4]{TCS Research, New York, USA}
\address[5]{Applied AI Research Lab, Department of Computer Science, University of South Dakota, USA}


\begin{abstract}
Continuous Latent Space (CLS) and Discrete Latent Space (DLS) models, like AttnUNet and VQUNet, have excelled in medical image segmentation. In contrast, Synergistic Continuous and Discrete Latent Space (CDLS) models show promise in handling fine and coarse-grained information. However, they struggle with modeling long-range dependencies. CLS or CDLS-based models, such as TransUNet or SynergyNet are adept at capturing long-range dependencies. Since they rely heavily on feature pooling or aggregation using self-attention, they may capture dependencies among redundant regions. This hinders comprehension of anatomical structure content, poses challenges in modeling intra-class and inter-class dependencies, increases false negatives and compromises generalization. Addressing these issues, we propose L2GNet, which learns global dependencies by relating discrete codes obtained from DLS using optimal transport and aligning codes on a trainable reference. L2GNet achieves discriminative on-the-fly representation learning without an additional weight matrix in self-attention models, making it computationally efficient for medical applications. Extensive experiments on multi-organ segmentation and cardiac datasets demonstrate L2GNet's superiority over state-of-the-art methods, including the CDLS method SynergyNet, offering an novel approach to enhance deep learning models' performance in medical image analysis.
\end{abstract}

\begin{keyword}
\KWD Cirrhotic liver segmentation \sep Abdominal MRI dataset  \sep liver segmentation \sep liver disease diagnosis \sep Liver Cancer \sep Abdominal Organ segmentation \sep T1-weighted MRI dataset \sep T2-weighted MRI dataset
\sep liver disease diagnosis \sep medical image segmentation \sep Transformer \sep Deep learning
\end{keyword}

\end{frontmatter}


\section{Introduction}
As reliance on medical image analysis grows in radiology rooms, the demand for precise, robust medical image segmentation techniques rises~\cite{meyer2018survey}. 
Deep learning has greatly improved our ability for medical image segmentation. Existing works in the literature have shown that medical image analysis tasks such as segmentation often require learning not just global features but also local features to capture region of interest (ROI) more precisely~\cite{chen2021TransUNet,synergynet,yan2019learning,chen2019collaborative,chaitanya2020contrastive}. Recently, continuous learning-based models (CLS), discrete learning-based models (DLS), and synergized continuous-discrete learning-based models (CDLS) have regained popularity and demonstrated significant success in capturing both global features, such as organ shapes, and local features, such as boundaries~\cite{chen2021TransUNet,santhirasekaram2022vector,synergynet,gorade2024harmonized}. However, there are inherent challenges and trade-offs associated with each approach.

\begin{figure*}[!t]\centering
\def\svgwidth{\columnwidth}
\includegraphics[width=\linewidth]{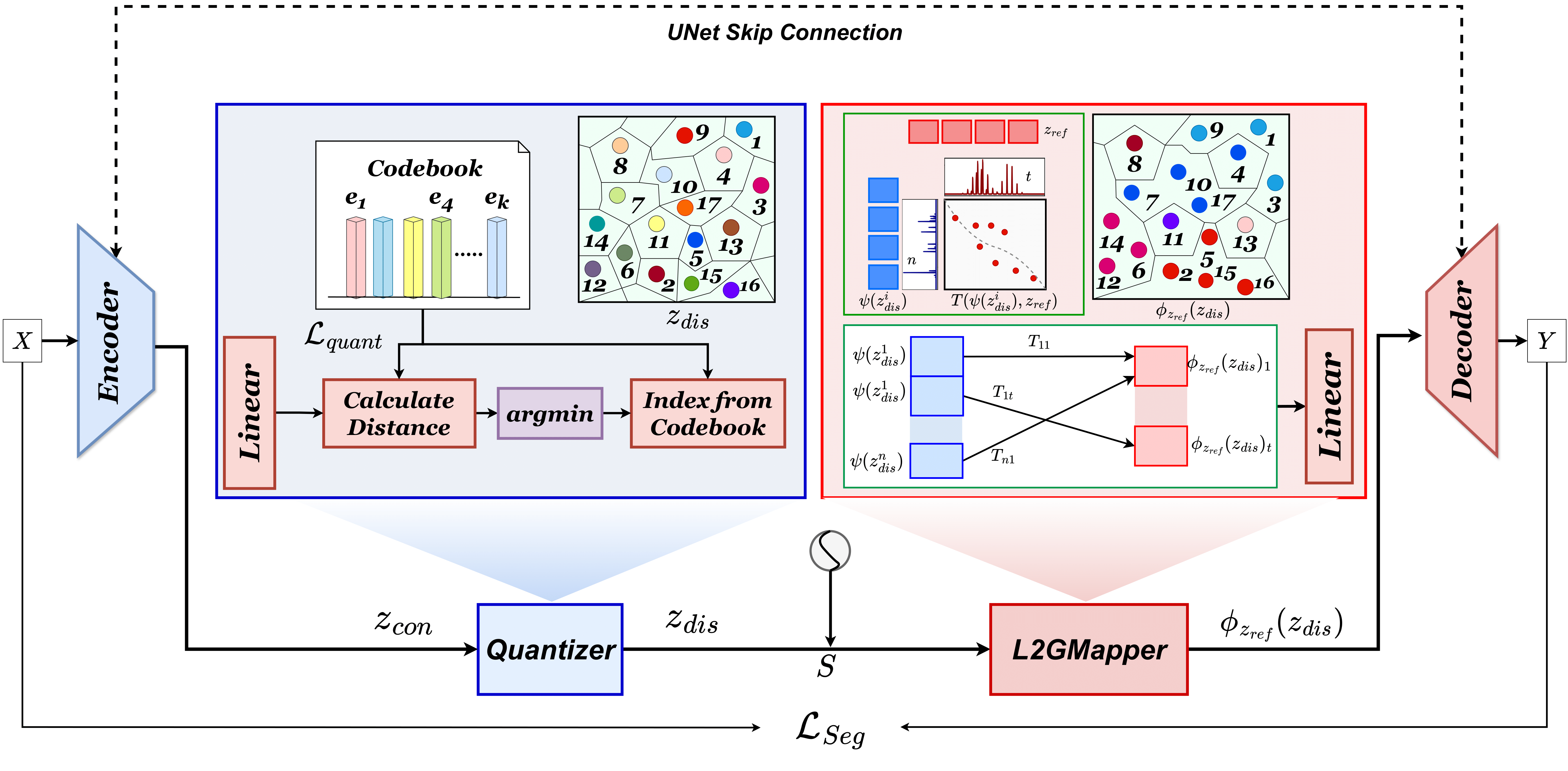} 
\caption{Illustrates the workflow of Proposed L2GNet.}
\label{fig:main_arch}
\end{figure*}

DLS methods excel in capturing structured local information and some global features, thanks to the quantization clustering effect introduced by vector quantization~\cite{van2017neural}. However, these methods often struggle with modeling long-range dependencies. On the other hand, CLS or CDLS-based models, such as TransUNet~\cite{chen2021TransUNet} or SynergyNet~\cite{synergynet}, are adept at capturing long-range dependencies. Still, they heavily rely on feature pooling or aggregation, typically using similarity weights for self-alignment~\cite{vaswani2017attention,dosovitskiy2020image}. This reliance on self-alignment mechanisms in CLS and CDLS methods has certain drawbacks. It tends to capture long-range dependencies between redundant regions rather than focusing on pertinent ones~\cite{gorade2024harmonized}. Consequently, these models may struggle to fully understand the content and positional arrangement of anatomical structures, leading to challenges in effectively modeling both intra-class and inter-class dependencies. This leads to higher false negatives and compromises the generalization capabilities of the model~\cite{khosla2020supervised,fu2020domain,gorade2024harmonized}. 
To address the challenges mentioned above, we propose a novel approach called \textit{L2GNet}. In contrast to the dot-product-based self-alignment, our approach anatomically pools together similar structures if they align well with the same part of a learnable reference. L2GNet comprises an encoder, quantizer, L2GMapper and decoder. The encoder is responsible for extracting a detailed continuous representation, and the quantizer module subsequently maps this representation to a compact discrete form (or codes) through vector quantization. This dimensionality reduction enables an efficient, structured local representation of anatomy while retaining crucial global information. The L2GMapper acts as a bridge, facilitating the learning of a global representation while capturing long-range dependencies between pertinent regions.

Drawing inspiration from optimal transport theory\cite{peyre2017computational,mairal2016end,mialon2020trainable,chen2019biological}, we achieve this by initially mapping codes to a reproducing kernel Hilbert space(RKHS) while preserving their positional information. Optimal transport facilitates the alignment of codes on a trainable reference using Sinkhorn distances\cite{cuturi2013sinkhorn}. This allows us to perform a weighted pooling operation, with weights determined by the transport plan between the codes and a learnable reference tailored to the specific segmentation task.

\textbf{Contributions}: (1) We introduce L2GNet, a novel architecture designed for learning local-to-global representations of anatomical structures while preserving long-range dependencies between pertinent regions. (2) To the best of our knowledge, discrete optimal transport kernel is never used as a lightweight alternative to existing attention mechanism-based bottlenecks in medical applications, which we propose herein. (3) We evaluate L2GNet on two segmentation benchmarks, including the Synapse~\cite{synapse} and ACDC~\cite{acdc}. Our proposed L2GNet consistently outperforms other methods based on CLS, DLS, and CDLS across all datasets. This confirms that L2GNet exhibits annotation efficiency and generalization capabilities. (4) Qualitative analyses affirm the effectiveness of L2GNet in capturing anatomical structures across different scales, showcasing its robust performance in handling both inter-class and intra-class variations.

\section{Preliminaries}

To establish the foundation for our novel approach, we first delve into optimal transport and vector quantization. These techniques play a critical role in our proposed method.

\textbf{Optimal Transport.}
\label{prelim:OT}
The optimal transport problem aims to find the most cost-effective way to transport mass from one distribution to another. Given two discrete measures represented by weights \(a\) and \(b\) on locations \(z\) and \(z'\), respectively, and pairwise costs \(C\), the entropic regularized Kantorovich relaxation~\cite{peyre2017computational} of OT is formulated as:
\begin{equation}
    \min_{T \in U(a, b)} \sum_{ij} M_{ij}T_{ij} - \varepsilon H(T),
\end{equation}
where \(H(T)\) is the entropic regularization term, \(\varepsilon\) controls sparsity, and \(U(a, b)\) is the space of admissible couplings. The problem is often solved using Sinkhorn's algorithm~\cite{cuturi2013sinkhorn}. In practice, when considering the evenly distributed mass, \(a\) and \(b\) become uniform measures. The resulting transport plan \(T\) provides information on how to distribute the mass from \(z\) to \(z'\) with minimal cost, allowing the alignment of features in a given code with a learned reference.

\textbf{Vector Quantization.} 
\label{prelim:VQ}
In VQVAE, Vector Quantization (VQ)~\cite{van2017neural} transforms continuous latent vectors $z_{con} \in \mathbb{R}^{dim}$ into discrete codes $e_k$ from the codebook $E \in \mathbb{R}^{K \times dim}$. The VQ process aims to find the code $e_k$ minimizing Euclidean distance to $z_{con}$, serving as the discrete representation $z_{dis}$. Training involves learning the codebook $E$ and mappings, minimizing quantization loss: 
\begin{equation}
    \mathcal{L}_{quant} = \lVert z_{con} - e_k \rVert_2^2.
\end{equation}
This enables dynamic learning of structured local information via discrete codes, preserving global information. We leverage evolving codes aligned with learnable references to capture global representations and dependencies between codes.

\textbf{Motivation.} The dot-product self-attention in transformers~\cite{vaswani2017attention} calculates the attention of the element using a dot product of linear transformations. However, it has drawbacks, including the need to store a large attention matrix \(W\) of size \(O(n^2)\). Moreover, learning only long-range dependencies through random patch interaction is insufficient for generalization. The proposed L2Gmapper module in L2GNet addresses this by focusing on learning a local-to-global representation of anatomical structures. It aligns codes with learnable references, preserving long-range dependencies between pertinent anatomical regions while reducing the size of the attention matrix \(W\) from quadratic to linear in length. This makes L2GNet memory efficient for medical image segmentation.


\section{Proposed Method}


\vspace{-1mm}
Given \(z_{\text{con}}\) obtained from the encoder, processed through the quantizer (sec.~\ref{prelim:VQ}) to derive discrete codes \(z_{dis}\), and a learned reference \(z_{ref}\) in the space \(X\) with \(t\) codes, we define an embedding \(\phi_{z_{ref}}(z_{dis})\) involving: (i) initial embedding of the codes of \(z_{dis}\) and \(z_{ref}\) to an RKHS \(\mathcal{H}\); (ii) alignment of \(z_{dis}\) codes to \(z_{ref}\) codes using optimal transport; (iii) weighted linear pooling of \(z_{dis}\) codes into \(t\) bins, resulting in an embedding \(\phi_{z_{ref}}(z_{dis})\) in \(\mathcal{H}^{t}\), as depicted in Fig. \ref{fig:main_arch}.

Let \(k\) represent the positive definite kernel with RKHS \(H\) and \(\psi:\mathbb{R}^d \rightarrow \mathcal{H}\) as its associated kernel embedding. The matrix \(k\) is of size \(n \times t\) and contains the comparisons \(k(z^i_{dis}, z^{i}_{ref})\) prior to alignment. In practical scenarios, where \(z_{dis}\) has finite dimensionality, it becomes computationally feasible to explicitly calculate the embedding \(\phi_{z_{ref}}(z_{dis})\). This is particularly advantageous for handling large-scale datasets, enabling the application of our method to supervised learning tasks. In scenarios where the discrete codes $z_{\text{dis}}$ are large, one can opt for an approximation using the Nystrom method \cite{williams2000using}, resulting in an embedding $\psi: \mathbb{R}^d \rightarrow \mathbb{R}^k$. The Nystrom method involves projecting points from the reproducing Hilbert space of the kernel ($\mathcal{H}$) onto a linear subspace $F$ with anchor points $k$, denoted as $F = \text{Span}(\phi(w_1), \ldots, \phi(w_k))$. The resulting embedding is explicitly formulated as $\psi(z^i_{\text{dis}}) = \frac{1}{\sqrt{\kappa(w, w)}}\kappa(w, z^i_{\text{dis}})$, where $\kappa(w, w)$ represents the $k \times k$ Gram matrix of $\kappa$ calculated at the anchor points $w = \{w_1, \ldots, w_k\}$, and $\kappa(w, z^i_{\text{dis}})$ belongs to $\mathbb{R}^k$. Parameters $w$ can be learned by propagating back in the context of a supervised task~\cite{mairal2016end}. This approach is particularly effective for tasks that involve high-dimensional images, such as MRIs. Implementing this approximation within our framework involves (i) substituting $k$ with a linear kernel, and (ii) replacing each element $z^i_{\text{dis}}$ with its embedding $\psi(z^i_{\text{dis}})$ in $\mathbb{R}^k$, considering a reference set with elements in $\mathbb{R}^k$. Subsequently, the transport plan between $z_{\text{dis}}$ and $z_{\text{ref}}$, denoted by the matrix $n \times t$ $T(\psi(z_{\text{dis}}), z_{\text{ref}})$, is defined as the unique solution of~\ref{prelim:OT} when choosing the cost $M = - \psi(z_{\text{dis}})$, and our embedding $\phi_{z_{\text{ref}}}(z_{\text{dis}})$ is defined as

\begin{equation}
\begin{split}
    \sqrt{t} \times \Bigg( 
    \sum_{i=1}^{n} T(\psi(z_{dis}), z_{ref})_i^1 \psi(z^{i}_{dis}) \times \mathcal{S}, \ldots, \\
    \sum_{i=1}^{n} T(\psi(z_{dis}), z_{ref})_{it}^t \psi(z^{i}_{dis}) \times \mathcal{S} 
    \Bigg)^\top
\end{split}
\end{equation}

Here, $t$ is the number of elements in $z_{\text{ref}}$. $\mathcal{S}$ allows us to consider the position of the codes, inspired by~\cite{mialon2020trainable}. We element-wise multiply $T(\psi(z_{\text{dis}}), z_{\text{ref}})$ by a distance matrix $S$ defined as $S_{ij} = e^{-\frac{1}{{\sigma^2}_{\text{pos}}} \left(\frac{i}{n} - \frac{j}{t}\right)^2}$. This accounts for similarity weights based on both content and position, crucial for tasks such as segmentation. The elements of $z_{\text{ref}}$ are non-linearly embedded and then aggregated in buckets, one for each element in the reference $z_{\text{ref}}$, given the values of $T(\psi(z_{\text{dis}}), z_{\text{ref}})$. $T(\psi(z_{\text{ref}}), z_{\text{ref}})$ is computed using Sinkhorn’s algorithm~\cite{cuturi2013sinkhorn}, easily adaptable to batches of samples $\psi(z_{\text{dis}})$ with varying lengths, enabling GPU-friendly forward computations of the embedding $\phi_{z_{\text{ref}}}$. Importantly, all Sinkhorn's operations are differentiable, allowing $z_{\text{ref}}$ to be optimized with stochastic gradient descent through backpropagation~\cite{mairal2016end}. Self-attention utilizes multiple heads to attend to different parts of the input. Similarly, to enhance the approximation of the transport plan, we reconstruct $z_{\text{ref}}$ with various references $z_{\text{ref}}^1, z_{\text{ref}}^2, \ldots, z_{\text{ref}}^q$. Specifically,
\[ \Phi_{z^1_{ref},\ldots,z^q_{ref}}(x) = \frac{1}{\sqrt{q}} (\Phi_{z^1_{ref}}(x), \ldots, \Phi_{z^q_{ref}}(x)), \]
where $q$ is the number of references (the factor $\frac{1}{\sqrt{q}}$ comes from the mean). Finally, we pass $\phi_{z_{\text{ref}}}(z_{\text{dis}})$ to the decoder to obtain the prediction $\hat{y}$. The outer optimization process employs a loss function comprising Binary Cross Entropy (BCE) and Dice similarity coefficient.

\vspace{-1.5mm}
\begin{equation}
\label{eq:1}
\mathcal{L}_{seg} = ({BCE}(y, \hat{y}) + (1 - {Dice}(y, \hat{y}))) +  \mathcal{L}_{quant},
\end{equation}
where ${BCE}(y, \hat{y})$ calculates binary cross entropy loss between predicted labels $y$ and ground truth segmentation $\hat{y}$, and ${Dice}(y, \hat{y})$ computes the dice similarity coefficient between $y$ and $\hat{y}$. $\mathcal{L}_{quant}$ is quantization loss as discussed in section-\ref{prelim:VQ}.

\section{Experiments and Results}

\begin{table*}[t]
\centering
\caption{Multiorgan Segmentation Results on Synapse Dataset. * denotes SynergyNet with total 10 heads(8-2). \textcolor{red}{Red} - best, \textcolor{blue}{Blue} - second-best.} \label{tab:mainResults}
\adjustbox{width=0.9\linewidth}{
\begin{tabular}{p{2.3cm} cccccccccc}
\toprule
& \multicolumn{8}{c}{\textbf{Synapse}} & & \\ 
\cmidrule(lr){1-11} 
\textbf{Method} & \multicolumn{2}{c}{\textbf{Mean}} & \multicolumn{8}{c}{\textbf{Class-wise Dice Similarity Coefficient (DSC)}}   \\ 
\cmidrule(lr){2-3} \cmidrule(lr){4-11}  
 & \textbf{DSC($\uparrow$)} & \textbf{HD($\downarrow$)} & \textbf{Aorta} & \textbf{GB} & \textbf{KL} & \textbf{KR} & \textbf{Liver} & \textbf{PC} & \textbf{SP} & \textbf{SM}   \\ 
 
\midrule
UNet & 77.54 & 38.26 & 85.52 & {61.86} & 80.57 & 77.24 & 94.37 & 54.72 & {87.95} & 78.12   \\

AttnUNet     & 75.57 & 36.97  & {55.92}  & 63.91  & 79.20  & 72.71  & 93.56  & 49.37  & 87.19  & 74.95  \\

TransUNet  &  {77.48} & 30.45  & 87.23  & {63.13}  & 81.87  & 77.02  & 94.08  & 55.86  & 85.08  & 75.62    \\

VQUNet  & {63.44}  & 68.79 & 78.99  & 50.74  & {67.32}  & 61.91  & {89.94}  & {33.96}  & {73.83}  & {50.87}   \\
SSNet & {78.36} & 32.48 &{86.42} & {61.16} & \textcolor{blue}{83.55} & {79.64} & {94.44} & {57.69} & {85.67} & {78.32}   \\

TranSSNet & {78.74} & 33.63 & {85.79} & {63.61} & {82.73} & {77.38} & {94.90} & {59.09} & {86.44} & {80.00}   \\

SynergyNet*  &  \textcolor{blue}{79.65}  & \textcolor{blue}{23.59} & {86.10}  & \textcolor{blue}{65.49}  & {82.78}  & {79.23}  & \textcolor{red}{95.06}  & {58.28}  & {88.95}  & \textcolor{blue}{{81.30}}   \\
\midrule
L2GNet(2-ref)  & 78.98 & 23.90 & \textcolor{red}{87.75} & 57.75 & 83.48 & 75.79 & 94.43  & \textcolor{blue}{64.85}  & \textcolor{blue}{90.43} & 77.33 \\

L2GNet(4-ref)  & \textcolor{red}{82.23} & \textcolor{red}{14.17} & 86.87 & \textcolor{red}{69.06} & \textcolor{red}{86.32} & \textcolor{red}{79.54} & \textcolor{red}{95.06}  & \textcolor{red}{67.49}  & \textcolor{red}{91.80}  & \textcolor{red}{81.77} \\

\bottomrule
\end{tabular}
}
\end{table*}

\textbf{Dataset and Experiment Settings.}
We conducted experiments on the Synapse dataset~\cite{synapse} for multi-organ segmentation and the ACDC~\cite{acdc} dataset for cardiac segmentation, following the same preprocessing and training configuration described in SynergyNet\cite{synergynet}. L2GNet employs a pre-trained ResNet50 encoder from ImageNet, and the quantizer module uses a codebook size of K = 512 with a hidden dimension of dim = 1024. Two L2GNet variants are evaluated; for instance, L2GNet(4-ref) indicates $q=4$ references (equivalent to heads in ViT) in the L2GMapper module. We set the number of iterations to 10 for Sinkhorn inner optimization. Both pre- and  post-bottleneck blocks consist of 2 convolution blocks, and the decoder matches the depth of the encoder.
We compare L2GNet against four CLS methods (UNet, Att-UNet~\cite{ronneberger2015u}, TransUNet~\cite{chen2021TransUNet}, SSNet~\cite{gorade2024harmonized}, TranSSNet~\cite{gorade2024harmonized}), one DLS method (VQUNet~\cite{santhirasekaram2022vector}), and one CDLS method (SynergyNet~\cite{synergynet}).

\begin{table*}[!t]
\centering
\caption{Cardiac segmentation results on ACDC dataset under Supervised Setting.}
\label{tab:mainResults2}

\adjustbox{max width=\linewidth}{
\begin{tabular}{p{2.1cm} lc c c c c c c} 
\toprule
\multirow{ 2}{*}{Method} & \multicolumn{8}{c}{\textbf{Supervised Setting}} \\ 
\cmidrule{2-9}
 & \multicolumn{2}{c}{\textbf{Mean}} & \multicolumn{3}{c}{\textbf{Class-wise DSC}} & \multicolumn{3}{c}{\textbf{Class-wise HD}} \\
\cmidrule(lr){2-3} \cmidrule(lr){4-6} \cmidrule(lr){7-9}
\textbf{} &  \textbf{DSC} &  \textbf{HD} & \textbf{RV} & \textbf{Myo} & \textbf{LV} & \textbf{RV} & \textbf{Myo} & \textbf{LV} \\
\midrule
UNet     & 87.94  &    1.98    & 84.62  & 84.52  & 93.68 & 3.81 & 1.10 & {1.05} \\
AttnUNet   & 86.90  &    2.10    & 83.27  & 84.33  & 93.53 & 3.84 & 1.14 & {1.11} \\
TransUNet     & {89.71}  &    {1.82}    & 86.67  &  {87.27}  &  {95.18} & 3.39 & 1.06 & \textcolor{blue}{1.04} \\
VQUNet            &     78.15    &    3.16     &    70.14 &    74.13  &  90.13  & 5.09 & 2.15 & 2.24  \\
SSNet     & {89.69} & 1.54 & {87.90} & {86.62} & {94.74}  & {2.49} & {1.07} & 1.08 \\
TranSSNet      & {91.32} & 1.30 & \textcolor{blue}{90.09} & {88.34} & \textcolor{blue}{95.53} & {1.85} & {1.02} & \textcolor{blue}{1.04} \\
SynergyNet      &  {89.78}  &  {1.86}  &  {87.68}  & {86.60}       &   {95.06} & 2.76 & 1.53 & 1.15 \\
\midrule
L2GNet(2-ref)      &  \textcolor{blue}{91.36}  &  \textcolor{red}{1.16}  &  \textcolor{red}{90.04}  & \textcolor{blue}{88.62} &  \textcolor{red}{95.55}    &  \textcolor{red}{1.42} & 1.02  &  \textcolor{red}{1.03}  \\
L2GNet(4-ref)      &  \textcolor{red}{91.44}  &  \textcolor{blue}{1.24}  &  89.97  & \textcolor{red}{88.84} &  95.50      & \textcolor{blue}{1.64} &  \textcolor{red}{1.01}  &  1.10  \\
\bottomrule
\end{tabular}}
\end{table*}

\begin{figure*}[t]
\centering
\def\svgwidth{\columnwidth}
\includegraphics[width=\linewidth]{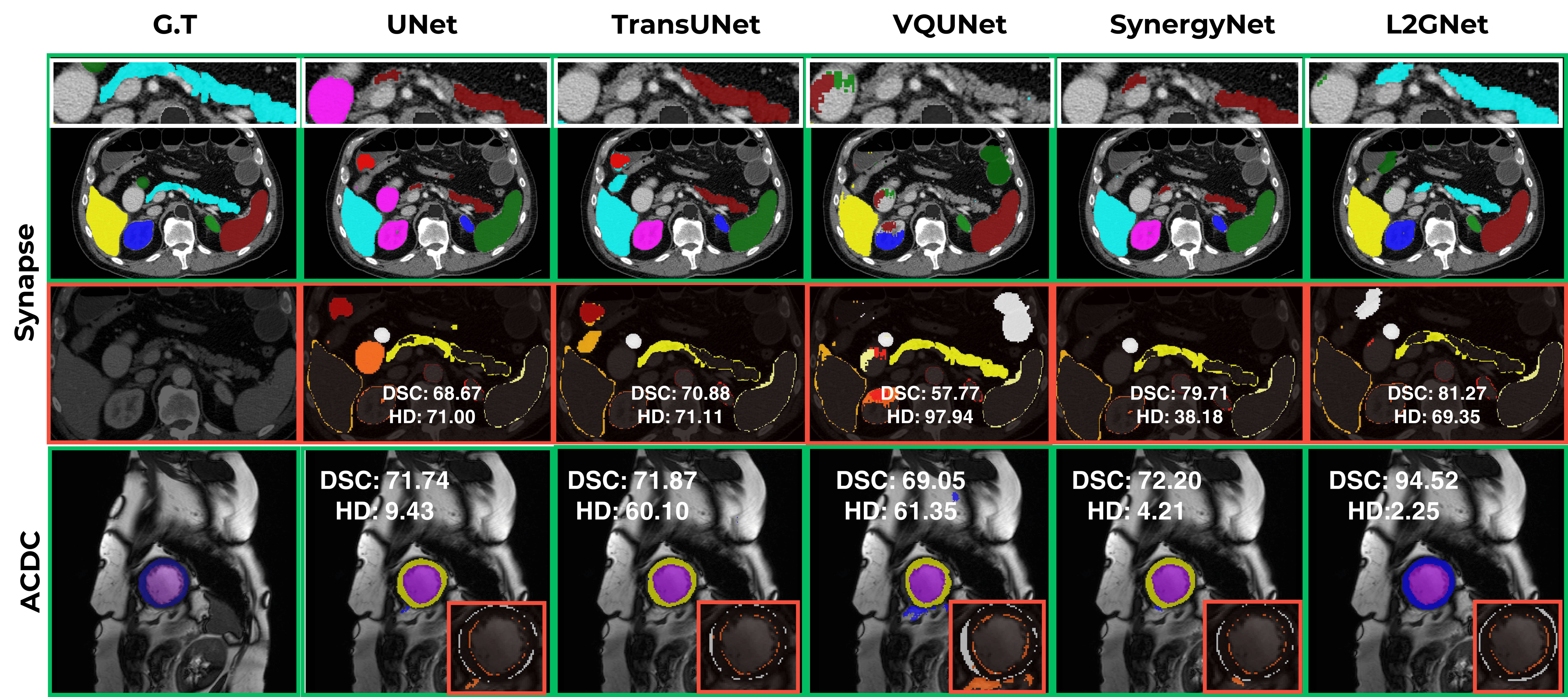}
\caption{Segmentation maps on Synapse and ACDC  datasets are shown with color-code (First three rows,
yellow: liver, blue: right kidney, green: left kidney, light blue: pancreas. Last row, blue, purple, and yellow represent the RV, LV, and
MYO, respectively.)
}
\label{qual:qual_res}
\end{figure*}

\textbf{Main Results.} In Table \ref{tab:mainResults}, L2GNet (4 refs), using only four references in its implementation, outperforms TransUNet and SynergyNet, which leverage 8 and 10 heads, respectively. While SynergyNet excels in delineating fine and coarse anatomical structures due to its CDLS nature, it still struggles with complex organs like pancreas. Similar limitations are observed in other CLS and DLS methods. In contrast, L2GNet accurately delineates anatomical structures at varying scales while maintaining fine boundaries and surpasses SynergyNet in delineating fine structures like the Aorta, gallbladder, kidney, spleen, etc. This underscores the contribution of the L2GMapper module in modeling long-range dependencies in pertinent regions, distinguishing it from other attention-based models. Furthermore, L2GNet effectively mitigates issues arising from dependencies between different classes in the segmentation process, reducing the risk of false negatives. As shown in Fig~\ref{qual:qual_res}, comparison methods misclassify the liver as the pancreas, highlighting their failure in learning inter-organ dependencies. Table~\ref{tab:mainResults2} presents L2GNet results on the ACDC cardiac segmentation task, demonstrating improved performance compared to other methods. In Fig~\ref{qual:qual_res}(last row), SynergyNet misclassifies the right ventricle (RV) as the myocardium (MYO), highlighting the significance of learning anatomical relations. .


\begin{table*}[!ht]
\label{tab:ablation}
\centering
\begin{minipage}{0.5\textwidth}
\centering
\caption{Codebook embedding size analysis.}\label{tab:codebook_analysis_1}
\adjustbox{max width=\linewidth}{
\begin{tabular}{cc|c c | c c|c c | c c}
\toprule
\multirow{2}{*}{$K_{dim}$}  & &\multicolumn{4}{c} {\textbf{Synapse}} & \multicolumn{4}{c}{\textbf{ACDC}} \\
\cmidrule(lr){3-6} \cmidrule(lr){7-10}
& & \multicolumn{2}{c}{\textbf{SynergyNet}} & \multicolumn{2}{c}{\textbf{L2GNet}} & 
\multicolumn{2}{c}{\textbf{SynergyNet}} & \multicolumn{2}{c}{\textbf{L2GNet}} \\
\cmidrule(lr){3-4} \cmidrule(lr){5-6} \cmidrule(lr){7-8} \cmidrule(lr){9-10}
& & \textbf{DSC} & \textbf{HD} & \textbf{DSC} & \textbf{HD} & \textbf{DSC} & \textbf{HD} & \textbf{DSC} & \textbf{HD} \\
\midrule
1024  & 

& {77.61} 
& {29.53} 
& 80.27
& 24.00
& {88.64} 
& 2.12
& \textcolor{blue}{91.41}
& 1.36
\\
512  & 
& 79.61 
& 23.89  
& \textcolor{red}{82.23}
& \textcolor{red}{14.17}
& 88.89 
& 1.86
& \textcolor{red}{91.44}
& \textcolor{red}{1.24}
\\
256  & 
& 79.21 
& 30.07
& \textcolor{blue}{81.35} 
& \textcolor{blue}{22.92}
& 89.18 
& 2.29
& 90.25
& \textcolor{blue}{1.35}
\\
128  & 
& 78.98 
& 35.81 
& 79.88
& 23.45
& 88.87
& 2.60
& 89.77
& 1.67
\\
64  & 
&  {77.29} 
& {88.67}
& 79.24
& 26.63
& {88.79} 
& {1.98} 
& 89.30
& 2.01
\\
\midrule
0  & 
&  {77.48} 
& {30.45}
& 78.69
& 27.96
& {89.71} 
& {1.82} 
& 89.30
& 2.01
\\
\bottomrule
\end{tabular}
}
\end{minipage}\hfill
\begin{minipage}{0.42\linewidth}
\centering
\caption{Analysis of 
 $q$}\label{tab:codebook_analysis_2}
\adjustbox{max width=\linewidth}{
\begin{tabular}{cc|cc|cc}
\toprule
\multirow{2}{*}{$q$}  & &\multicolumn{2}{c} {\textbf{Synapse}} & \multicolumn{2}{c}{\textbf{ACDC}} \\
\cmidrule(lr){3-4} \cmidrule(lr){5-6}
& & \textbf{DSC} & \textbf{HD} & \textbf{DSC} & \textbf{HD} \\
\midrule
& \multicolumn{4}{c}{{\textit{SynergyNet}}} \\ 
\midrule
2-2  & &  78.81 & 26.19  & 88.96 & 2.41 \\
8-2  & &  \textcolor{blue}{79.65} & \textcolor{blue}{23.29} & {89.78} & 1.86 \\
8-8  & &  {77.33} & {20.56} & {89.68} & {2.14} \\
\midrule
& \multicolumn{4}{c}{{\textit{L2GNet}}} \\ 
\midrule
2  & & {78.98} & {23.90} & \textcolor{blue}{91.36} & \textcolor{red}{1.16} \\
4  & &  \textcolor{red}{82.23} &  \textcolor{red}{14.17} &  \textcolor{red}{91.44} & \textcolor{blue}{1.24}  \\
\bottomrule
\end{tabular}
}
\end{minipage}

\end{table*}


\begin{table*}[!t]
\centering
\caption{Analysis of $q$ for Synapse and ACDC datasets.}
\label{tab:codebook_analysis_2}
\adjustbox{max width=\linewidth}{
\begin{tabular}{c ccc c|cc ccc ccc c} 
\toprule
\multirow{2}{*}{$q$} & \multicolumn{3}{c}{\textbf{Synapse (SynergyNet)}} & \multicolumn{3}{c}{\textbf{ACDC (SynergyNet)}} & \multicolumn{3}{c}{\textbf{Synapse (L2GNet)}} & \multicolumn{3}{c}{\textbf{ACDC (L2GNet)}} & \multirow{2}{*}{$q$} \\
\cmidrule(lr){2-4} \cmidrule(lr){5-7} \cmidrule(lr){8-10} \cmidrule(lr){11-13}
& \textbf{DSC} & \textbf{HD} & & \textbf{DSC} & \textbf{HD} & & \textbf{DSC} & \textbf{HD} & & \textbf{DSC} & \textbf{HD} & & \\
\midrule
2-2 & 78.81 & 26.19 & & 88.96 & 2.41 & & 78.98 & 23.90 & & \textcolor{blue}{91.36} & \textcolor{red}{1.16} & & 2 \\
8-2 & \textcolor{blue}{79.65} & \textcolor{blue}{23.29} & & 89.78 & 1.86 & & \textcolor{red}{82.23} & \textcolor{red}{14.17} & & \textcolor{red}{91.44} & \textcolor{blue}{1.24} & & 4 \\
\bottomrule
\end{tabular}}
\end{table*}

\begin{table*}[!t]
\centering
\caption{Codebook embedding size analysis.}\label{tab:codebook_analysis_1}
\adjustbox{max width=\linewidth}{
\begin{tabular}{cc|c c | c c|c c | c c}
\toprule
\multirow{2}{*}{$K_{dim}$}  & &\multicolumn{4}{c} {\textbf{Synapse}} & \multicolumn{4}{c}{\textbf{ACDC}} \\
\cmidrule(lr){3-6} \cmidrule(lr){7-10}
& & \multicolumn{2}{c}{\textbf{SynergyNet}} & \multicolumn{2}{c}{\textbf{L2GNet}} & 
\multicolumn{2}{c}{\textbf{SynergyNet}} & \multicolumn{2}{c}{\textbf{L2GNet}} \\
\cmidrule(lr){3-4} \cmidrule(lr){5-6} \cmidrule(lr){7-8} \cmidrule(lr){9-10}
& & \textbf{DSC} & \textbf{HD} & \textbf{DSC} & \textbf{HD} & \textbf{DSC} & \textbf{HD} & \textbf{DSC} & \textbf{HD} \\
\midrule
1024  & 
& {77.61} & {29.53} & 80.27 & 24.00 & {88.64} & 2.12 & \textcolor{blue}{91.41} & 1.36 \\
512  & 
& 79.61 & 23.89  & \textcolor{red}{82.23} & \textcolor{red}{14.17} & 88.89 & 1.86 & \textcolor{red}{91.44} & \textcolor{red}{1.24} \\
256  & 
& 79.21 & 30.07 & \textcolor{blue}{81.35} & \textcolor{blue}{22.92} & 89.18 & 2.29 & 90.25 & \textcolor{blue}{1.35} \\
128  & 
& 78.98 & 35.81 & 79.88 & 23.45 & 88.87 & 2.60 & 89.77 & 1.67 \\
64  & 
&  {77.29} & {88.67} & 79.24 & 26.63 & {88.79} & {1.98} & 89.30 & 2.01 \\
\midrule
0  & 
&  {77.48} & {30.45} & 78.69 & 27.96 & {89.71} & {1.82} & 89.30 & 2.01 \\
\bottomrule
\end{tabular}
}
\end{table*}

\section{Conclusion}
We introduce L2GNet, a novel bottleneck architecture, excelling in capturing local-to-global long-range dependencies between pertinent regions with lower computational complexity compared to self-alignment-based bottlenecks like TransUNet and SynergyNet. Additionally, our results emphasize L2GNet's efficiency and interpretability, positioning it as a promising framework for medical segmentation. In the future, we plan to expand L2GNet by exploring 3D volumes, building upon our successful experiments with 2D slices. Additionally, we aim to enhance computational efficiency and generalization capabilities by integrating of foundational models by integrating it with L2GNet.

%





\begin{thebibliography}{1}

\bibitem{kawaguchi2017generalization} K. Kawaguchi, L. P. Kaelbling, and Y. Bengio, ``Generalization in deep learning,'' \textit{arXiv preprint arXiv:1710.05468}, vol. 1, no. 8, 2017.

\bibitem{meyer2018survey} P. Meyer, V. Noblet, C. Mazzara, and A. Lallement, ``Survey on deep learning for radiotherapy,'' \textit{Computers in Biology and Medicine}, vol. 98, pp. 126--146, 2018.

\bibitem{yan2019learning} Z. Yan, X. Han, C. Wang, Y. Qiu, Z. Xiong, and S. Cui, ``Learning mutually local-global U-nets for high-resolution retinal lesion segmentation in fundus images,'' in \textit{Proc. IEEE ISBI}, pp. 597--600, 2019.

\bibitem{chen2019collaborative} W. Chen, Z. Jiang, Z. Wang, K. Cui, and X. Qian, ``Collaborative global-local networks for memory-efficient segmentation of ultra-high resolution images,'' in \textit{Proc. IEEE/CVF CVPR}, pp. 8924--8933, 2019.

\bibitem{chaitanya2020contrastive} K. Chaitanya, E. Erdil, N. Karani, and E. Konukoglu, ``Contrastive learning of global and local features for medical image segmentation with limited annotations,'' \textit{Advances in Neural Information Processing Systems}, vol. 33, pp. 12546--12558, 2020.

\bibitem{vaswani2017attention} A. Vaswani et al., ``Attention is all you need,'' \textit{Advances in Neural Information Processing Systems}, vol. 30, 2017.

\bibitem{dosovitskiy2020image} A. Dosovitskiy et al., ``An image is worth 16x16 words: Transformers for image recognition at scale,'' \textit{arXiv preprint arXiv:2010.11929}, 2020.

\bibitem{liu2021swin} Z. Liu et al., ``Swin transformer: Hierarchical vision transformer using shifted windows,'' in \textit{Proc. IEEE/CVF ICCV}, pp. 10012--10022, 2021.

\bibitem{zhou2018unetpp} Z. Zhou, M. M. R. Siddiquee, N. Tajbakhsh, and J. Liang, ``UNet++: A nested U-Net architecture for medical image segmentation,'' in \textit{Proc. Int. Workshop Deep Learning Med. Image Analysis}, pp. 3--11, 2018.

\bibitem{oktay2018attention} O. Oktay et al., ``Attention U-Net: Learning where to look for the pancreas,'' \textit{arXiv preprint arXiv:1804.03999}, 2018.

\bibitem{cao2022swin} H. Cao et al., ``Swin-UNet: UNet-like pure transformer for medical image segmentation,'' in \textit{Proc. ECCV}, pp. 205--218, 2022.

\bibitem{chen2021TransUNet} J. Chen et al., ``TransUNet: Transformers make strong encoders for medical image segmentation,'' \textit{arXiv preprint arXiv:2102.04306}, 2021.

\bibitem{santhirasekaram2022vector} A. Santhirasekaram et al., ``Vector quantisation for robust segmentation,'' in \textit{Proc. MICCAI}, pp. 663--672, 2022.

\bibitem{van2017neural} A. Van Den Oord and O. Vinyals, ``Neural discrete representation learning,'' \textit{Advances in Neural Information Processing Systems}, vol. 30, 2017.

\bibitem{heidari2023hiformer} M. Heidari et al., ``HiFormer: Hierarchical multi-scale representations using transformers for medical image segmentation,'' in \textit{Proc. WACV}, pp. 6202--6212, 2023.

\bibitem{synergynet} V. Gorade, S. Mittal, D. Jha, and U. Bagci, ``SynergyNet: Bridging the gap between discrete and continuous representations for precise medical image segmentation,'' in \textit{Proc. WACV}, pp. 7768--7777, 2024.

\bibitem{gorade2024harmonized} V. Gorade, S. Mittal, D. Jha, R. Singhal, and U. Bagci, ``Harmonized spatial and spectral learning for robust and generalized medical image segmentation,'' \textit{arXiv preprint arXiv:2401.10373}, 2024.

\bibitem{khosla2020supervised} P. Khosla et al., ``Supervised contrastive learning,'' \textit{Advances in Neural Information Processing Systems}, vol. 33, pp. 18661--18673, 2020.

\bibitem{fu2020domain} S. Fu et al., ``Domain adaptive relational reasoning for 3D multi-organ segmentation,'' in \textit{Proc. MICCAI}, pp. 656--666, 2020.

\bibitem{ronneberger2015u} O. Ronneberger, P. Fischer, and T. Brox, ``U-Net: Convolutional networks for biomedical image segmentation,'' in \textit{Proc. MICCAI}, pp. 234--241, 2015.

\bibitem{cuturi2013sinkhorn} M. Cuturi, ``Sinkhorn distances: Lightspeed computation of optimal transport,'' \textit{Advances in Neural Information Processing Systems}, vol. 26, 2013.

\bibitem{mialon2020trainable} G. Mialon et al., ``A trainable optimal transport embedding for feature aggregation and its relationship to attention,'' \textit{arXiv preprint arXiv:2006.12065}, 2020.

\bibitem{williams2000using} C. Williams and M. Seeger, ``Using the Nyström method to speed up kernel machines,'' \textit{Advances in Neural Information Processing Systems}, vol. 13, 2000.

\bibitem{peyre2017computational} G. Peyré, M. Cuturi, et al., ``Computational optimal transport,'' \textit{Center for Research in Economics and Statistics Working Papers}, no. 2017-86, 2017.

\bibitem{mairal2016end} J. Mairal, ``End-to-end kernel learning with supervised convolutional kernel networks,'' \textit{Advances in Neural Information Processing Systems}, vol. 29, 2016.

\bibitem{chen2019biological} D. Chen, L. Jacob, and J. Mairal, ``Biological sequence modeling with convolutional kernel networks,'' \textit{Bioinformatics}, vol. 35, no. 18, pp. 3294--3302, 2019.

\bibitem{synapse} \textit{Multi-Atlas Abdomen Labeling Challenge: Synapse Multi-Organ Segmentation Dataset}, Synapse Consortium, 2015. [Online]. Available: \url{https://www.synapse.org/#!Synapse:syn3193805/wiki/217789}

\bibitem{acdc} \textit{ACDC (Automated Cardiac Diagnosis Challenge)}, 2017. [Online]. Available: \url{https://www.creatis.insa-lyon.fr/Challenge/acdc}

\bibitem{tarvainen2017mean} A. Tarvainen and H. Valpola, ``Mean teachers are better role models: Weight-averaged consistency targets improve semi-supervised deep learning results,'' \textit{Advances in Neural Information Processing Systems}, vol. 30, 2017.

\bibitem{vu2019advent} T.-H. Vu et al., ``Advent: Adversarial entropy minimization for domain adaptation in semantic segmentation,'' in \textit{Proc. IEEE/CVF CVPR}, pp. 2517--2526, 2019.

\bibitem{yu2019uncertainty} L. Yu et al., ``Uncertainty-aware self-ensembling model for semi-supervised 3D left atrium segmentation,'' in \textit{Proc. MICCAI}, pp. 605--613, 2019.

\bibitem{verma2022interpolation} V. Verma et al., ``Interpolation consistency training for semi-supervised learning,'' \textit{Neural Networks}, vol. 145, pp. 90--106, 2022.

\bibitem{chen2021semi} X. Chen et al., ``Semi-supervised semantic segmentation with cross pseudo supervision,'' in \textit{Proc. IEEE/CVF CVPR}, pp. 2613--2622, 2021.

\end{thebibliography}

\end{document}